\documentclass{article}
\usepackage{spconf,amsmath,graphicx}
\usepackage{xspace}
\makeatletter
\DeclareRobustCommand\onedot{\futurelet\@let@token\@onedot}
\def\@onedot{\ifx\@let@token.\else.\null\fi\xspace}

\def\etc{etc\onedot}

\usepackage{booktabs}
\usepackage{multirow}
\usepackage{comment}

\graphicspath{{figures/}}
\usepackage{xcolor}
\usepackage{mathtools}

\newcommand{\Eq}[1]{Eq.~(\ref{eq:#1})}
\newcommand{\eq}[1]{\Eq{#1}}
\newcommand{\fig}[1]{Fig~\ref{fig:#1}}

\usepackage{booktabs}

\usepackage{url}
\usepackage{xcolor}
\title{Recurrent knowledge distillation}
\name
{
	Silvia L. Pintea$^1$,
	Yue Liu$^2$,
	Jan C. van Gemert$^1$
}
\address
{$^1$Vision Lab, Delft University of Technology, Netherlands\\
$^2$
School of Electrical Engineering and Computer Science, KTH, Stockholm, Sweden
}
\begin{document}
\topmargin=0mm
\maketitle
\begin{abstract}
Knowledge distillation compacts deep networks by letting a small student network learn from a large teacher network. 
The accuracy of knowledge distillation recently benefited from adding residual layers.  
We propose to reduce the size of the student network even further by recasting multiple residual layers in the teacher network into a single recurrent student layer. 
We propose three variants of adding recurrent connections into the student network, and show experimentally 
on CIFAR-10, Scenes and MiniPlaces, that we can reduce the number of parameters at little loss in accuracy.
\end{abstract}

\section{Introduction}

Deep learning requires deep computational pockets. 
Current models for image classification~\cite{he2016deep,lin2013network}, object detection~\cite{dai2016r,lin2016feature}, 
semantic segmentation~\cite{he2017mask,shelhamer2017fully} use millions of parameters. 
The memory requirements of such large models prevent real-world applicability in limited memory scenarios such as surveillance, home safety devices, industrial robots, autonomous driving, \etc. 
In this paper we investigate low-memory approximations of high-accuracy models. 

\begin{figure}
	\centering
	\includegraphics[width=0.40\textwidth]{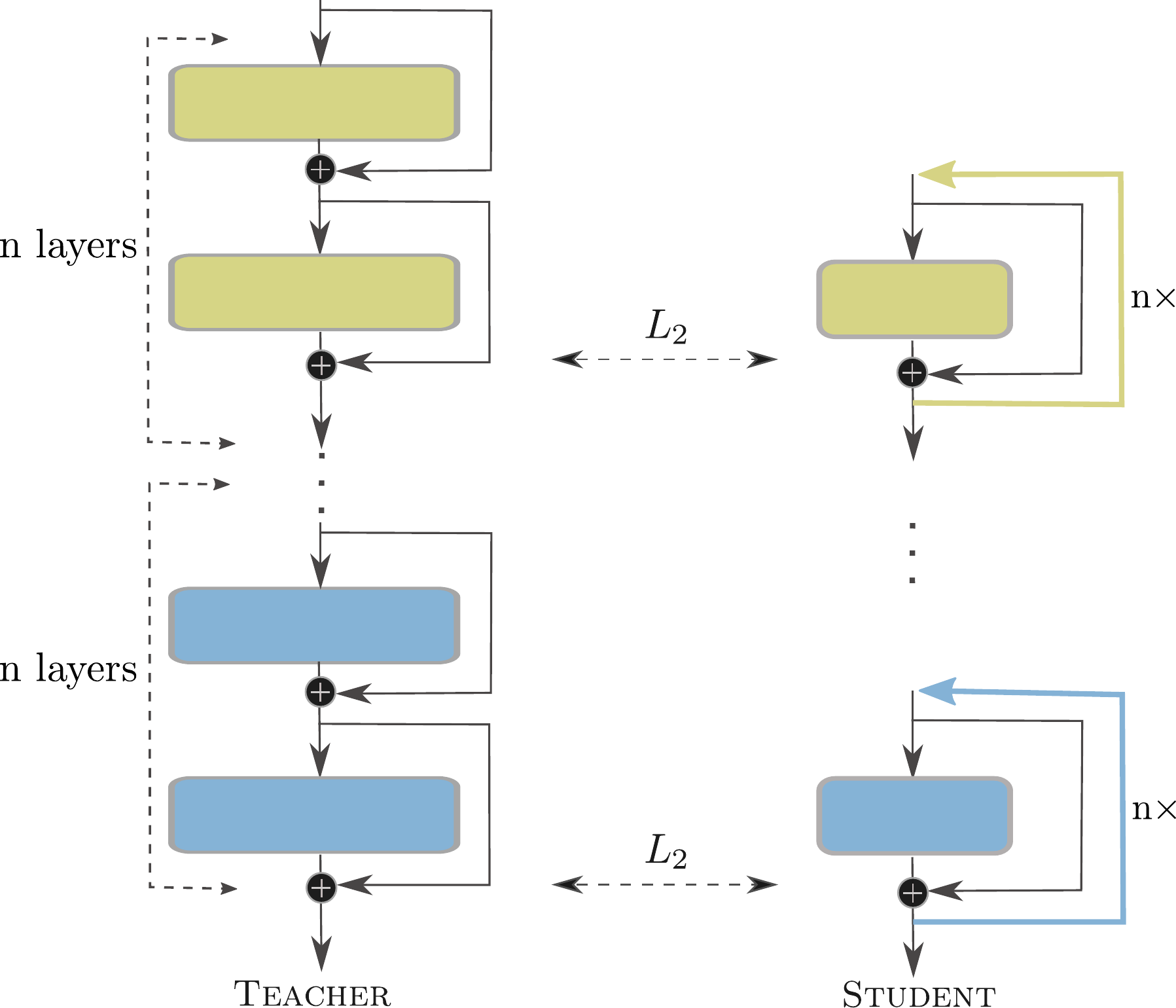}
	\caption{\small Multiple residual layers in the teacher network (left) are mapped to a single layer in the student network (right) through recurrence. The $L_2$ measures similarity between networks, and similar colors correspond to similar layers between networks. Note the reduction in layer parameters. 
	}
    \vspace{-15px}
	\label{fig:idea}
\end{figure}

Knowledge distillation~\cite{hinton2015distilling,ba2014deep} exploits large teacher models to help train student models that are more compact yet retain good accuracy. Recent methods use thin yet deep students~\cite{zagoruyko2017paying}, reaping the benefits of residual layers~\cite{he2016deep} to train extremely deep architectures.
For training such deep networks, the teacher-student similarity losses are computed at similar depths in the architecture~\cite{romero2015fitnets,zagoruyko2017paying}, requiring student models to have similar depths as their teachers.

In this paper we propose to further reduce the parameters of deep student networks by sharing weights between residual layers. We are inspired by residual learning seen as an iterative refinement scheme~\cite{greff2017highway}, and that multiple residual connections can be seen as an iterative recurrent connection ``unrolled" over time~\cite{liao31bridging,boulch2017sharesnet,veit2016residual}. 
We propose the use of recurrent connections for compacting the information of multiple residual layers of the teacher network into a single recurrent layer 
in the student model. In \fig{idea} we illustrate our approach. We make the following contributions: (i) student memory reduction by using recurrent connections instead of residuals;
(ii) exploring three variants of adding recurrence; 
(iii) experiments on CIFAR-10, Scenes and MiniPlaces show
substantially reduced parameters with similar accuracy.

\begin{figure*}[t]
	\centering
	\small
	\begin{tabular}{c@{\hskip 0.4in}c@{\hskip 0.4in}c@{\hskip 0.4in}c}
		\includegraphics[trim={0 14.4cm 0 0},clip,width=0.20\textwidth]{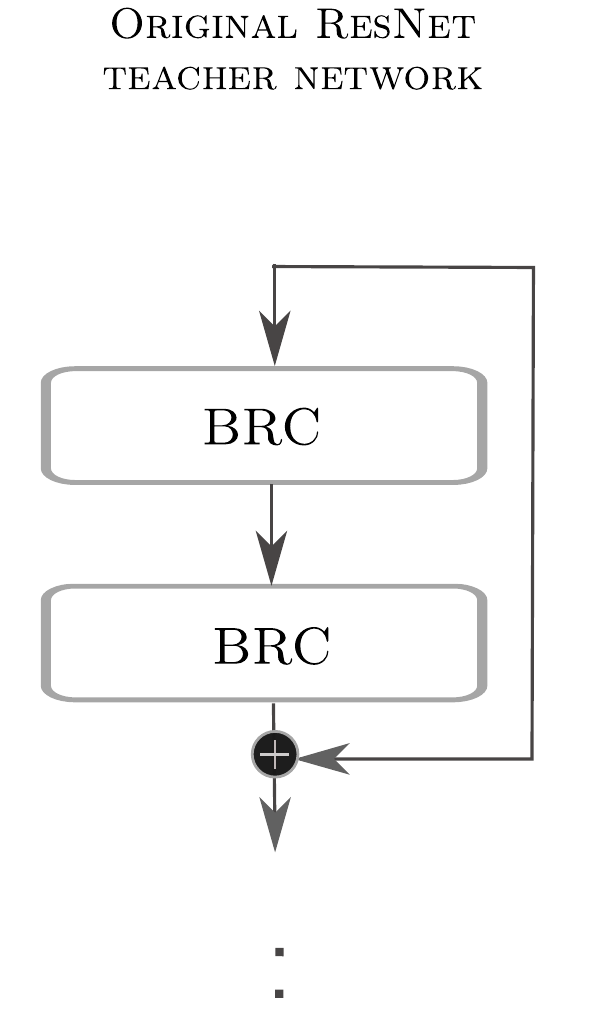} &
		\includegraphics[trim={0 14.4cm 0 0},clip,width=0.20\textwidth]{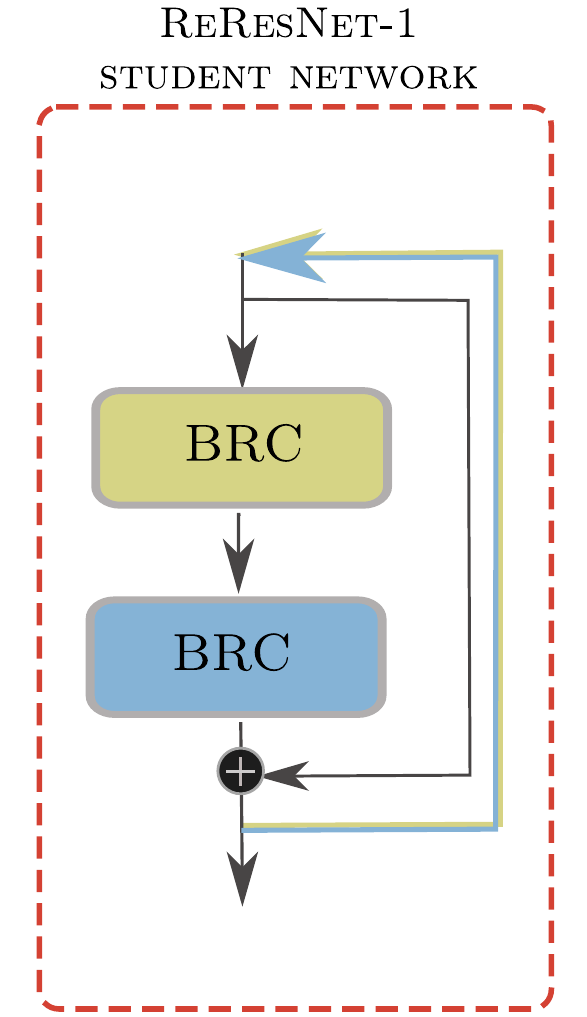} &
		\includegraphics[trim={0 14.4cm 0 0},clip,width=0.20\textwidth]{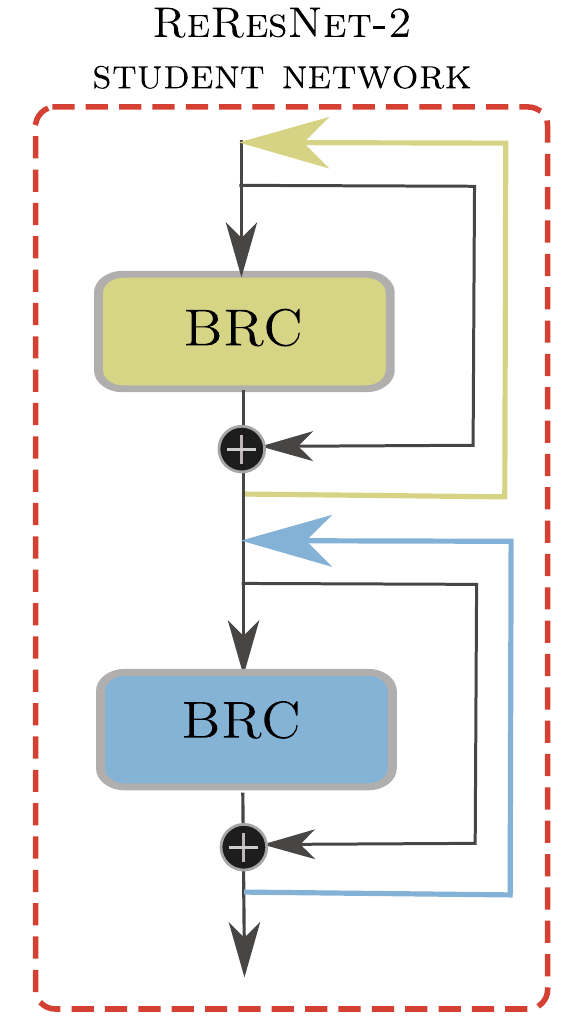} &
		\includegraphics[trim={0 14.4cm 0 0},clip,width=0.20\textwidth]{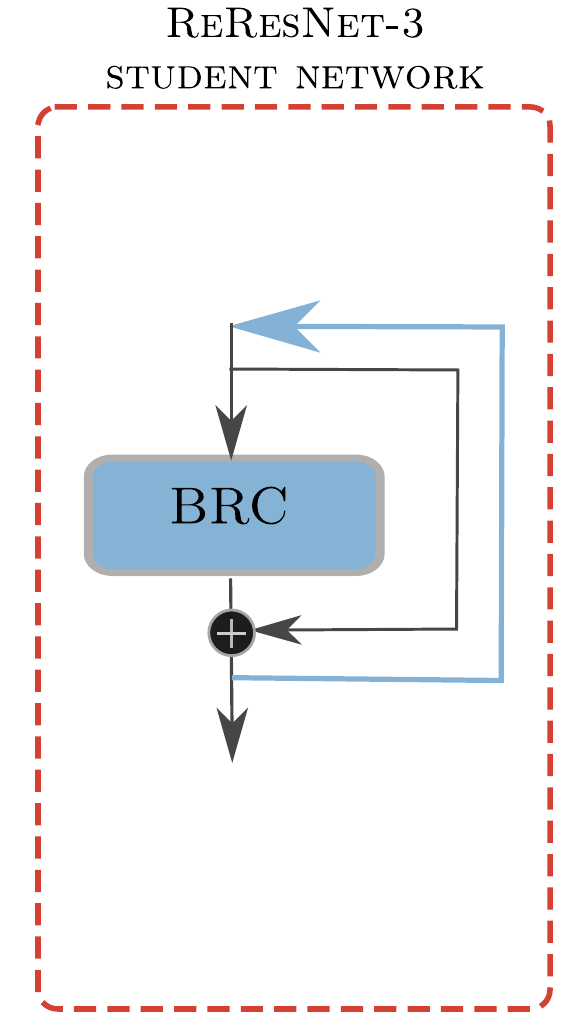} \\
		(a) Original. &
		(b) \textsc{ReResNet-1} & 
		(c) \textsc{ReResNet-2} & 
		(d) \textsc{ReResNet-3}\\
	\end{tabular}
	\caption{\small
		Our three proposed variants for the student models together with the original block.
		We display without color the units that are not tied over time and in the same color (yellow or blue) the units that share parameters over time.
		(a) The original residual blocks, each with two BRC (BN-RELU-Conv) units inside.
		(b) \textsc{ReResNet-1}: both units are tied over time in an interleaved fashion: the first one (in yellow) is tied at $t \in \{1 .., 2 n + 1\}$, and the second (in blue) at $t \in \{2 ..,  2 n\}$. 
		(c) \textsc{ReResNet-2}: the two BRC units broken apart, each being shared sequentially: the first one (in yellow) is tied at $t \in \{ 1 .., n \}$, and the second (in blue) 
		at $t \in \{n + 1 .., 2n\}$.
		(d) \textsc{ReResNet-3}: the two BRC units are collapsed into a single unit, shared at $t \in \{1 .. , n\}$.
		Where $n$ denotes the number of recurrence steps.	
	}
	\label{fig:reresnet}
\end{figure*}

\section{Related work}\label{sec:related}



\vspace{3px}\textbf{Knowledge distillation}. Knowledge between multiple modalities~\cite{gupta2016cross} can be transferred. 
Doing the opposite of our aim, \cite{chan2015transferring} trains a deep architecture from a recurrent language model,
while~\cite{chen2015net2net,yim2017gift} improve time and accuracy, rather than reducing parameters.
In contrast to these works, we aim at parameter reduction by training a low-parameter student network from a parameter-intensive residual teacher network 
through the use of recurrent connections. 

\vspace{3px} \noindent \textbf{Time efficiency.} Extremely deep networks can be trained efficiently by controlling the information flow through the use of multiple 
interconnections between layers and gating~\cite{srivastava2015highway,srivastava2015training}. While in \cite{lebedev2016fast} speedups are obtained by pruning convolutional kernels in a group-wise fashion.
The work in \cite{zagoruyko2016wide} uses shallower wider neural networks for training time efficiency. 
Unlike these methods, we do not focus on improving the speed, but on substantially reducing the number of parameters.


\vspace{3px}\noindent \textbf{Memory efficiency.} 
Network parameters can be compressed through low-rank and sparse decomposition of weight matrices~\cite{yu2017compressing,zhang2016accelerating}. Alternatively, binarization~\cite{rastegari2016xnor,daniely2017sketching} can reduce memory usage at the expense of a small drop in performance. Others exploit circular matrices in the Fourier domain for projecting the network feature maps for dimensionality reduction~\cite{cheng2015exploration,wang2017beyond}. Our method is intended to benefit before such approximations are applied. 
We train a low-parameter student network to mimic as closely as possible a parameter-intensive teacher network.

%
%

\section{Method}\label{sec:method}


\subsection{Recurrent ResNet for knowledge distillation}\label{ssec:reresnet}
Given a residual block $i$, with a corresponding mapping function, $f_i(\cdot)$, 
and weights $\mathbf{W}_i$, the output of its application, $\hat{f}_i(\cdot)$, to its input feature map, $\mathbf{x}_{i-1}$, is defined~\cite{he2016deep} as:
\begin{equation}
	\mathbf{x}_{i} = \hat{f}_{i} ( \mathbf{x}_{i-1}, \mathbf{W}_{i}) =  f_{i} ( \mathbf{x}_{i-1}, \mathbf{W}_{i}) + \mathbf{x}_{i-1}.
	\label{eq:resnet}
\end{equation}
All the feature maps in one residual block have the same width and height, therefore, the same spatial scale.
We define a recurrent unit at time $t > 1$, by sharing the weights at the same spatial scale in the residual block.
\begin{equation}
	\mathbf{x}_{i}(t) = \hat{f}_{i} (\mathbf{x}_i(t-1), \mathbf{W}_{i}),
	\label{eq:shared}
\end{equation}
where the feature maps $\mathbf{x}_i$, become a function of time, $t$.

Each residual block of a ResNet architecture \cite{he2016deep} consists of two BRC units, composed of: batch normalization (BN), rectified linear units (RELU), and a convolution (Conv).
We coin our recurrent residual model: \textsc{ReResNet}. 
Starting from a residual teacher network \cite{he2016deep}, we consider all three different possibilities of adding recurrence into the student network, 
by recurring both BRC units in a residual block.
\fig{reresnet} illustrates these three variants of the student model.

\vspace{3px}\noindent \textbf{ReResNet-1: interleaved parameter tying.} 
The first unit is tied over odd timesteps, $t = 2n+1$, and therefore its associated weights are shared over these timesteps. 
The second unit in the block is tied over even timesteps $t = 2n$, where $n$ is the number of recurrence steps considered. See \fig{reresnet}(b).

\vspace{3px}\noindent \textbf{ReResNet-2: sequential parameter tying. }
The first unit is tied over the timesteps $t = \{1, .., n\}$, and their weights are shared over those timesteps.
While the second unit is tied over timesteps $t = \{n+1, .., 2n\}$. See \fig{reresnet}(c).   

\vspace{3px}\noindent \textbf{ReResNet-3: single unit with shared weights.}
Collapsing the two BRC units to only one unit with shared weights which is then recurred over time $t = \{1, .., n\}$. See \fig{reresnet}(d).
\subsection{Incorporating the teacher-student similarity loss}\label{ssec:feature}


Our loss $\mathcal{L}(\mathbf{x}, \{\mathbf{W}^s\})$ over the input data $\mathbf{x}$ and network parameters of the student model, $\mathbf{W}^s$,
is a combination of the softmax classification loss, $\mathcal{L}_{\text{cls}}$, and the teacher-student similarity loss, $\mathcal{L}_{\text{ts}}$:
\begin{equation}
	\mathcal{L}(\mathbf{x}, \{\mathbf{W}^s\}) = \mathcal{L}_{\text{cls}}(\mathbf{x}, \{\mathbf{W}^s\}) 
	 + \lambda \mathcal{L}_{\text{ts}}(\mathbf{x}, \{\mathbf{W}^s, \mathbf{W}^t\}),
	\label{eq:loss}
\end{equation}
where the weights of the teacher model $\mathbf{W}^t$,  are fixed, and $\lambda$ is the trade off parameter between the two losses.
The teacher-student similarity loss $\mathcal{L}_{\text{ts}}$, is the $L_2$ distance between the teacher-student activations at a set of layer pairs, $\mathcal{K}$:
\scriptsize
\begin{equation}
	 \sum_{(i, j) \in \mathcal{K}} 
	 \left( \frac{ \sum_{c=1}^N \hat{f}_i^c(\mathbf{x}_i, \mathbf{W}_i^s)^2 }{ \Vert \sum_{c=1}^N \hat{f}_i^c(\mathbf{x}_i, \mathbf{W}_i^s)^2 \Vert_2 }  -  
		 \frac{ \sum_{c=1}^M \hat{f}_j^c(\mathbf{x}_j, \mathbf{W}_j^t)^2 }{ \Vert \sum_{c=1}^M \hat{f}_j^c(\mathbf{x}_j, \mathbf{W}_j^t)^2 \Vert_2 } \right)^2,
\end{equation}
\normalsize
where $N$ and $M$ are the number of channels in the 
student and teacher feature maps, respectively. 
To allow for different number of channels between the teacher network and the student network, the squared values of the feature maps in the two networks are first accumulated over the channels 
as in \cite{zagoruyko2017paying}, and subsequently, the feature maps are normalized.  
   
\subsection{Implementing recurrence and shared gradients}
We implement recurrence by sharing the parameters of the convolutional layers, 
allowing each depth to learn its own BN (Batch Normalization) parameters.
We do so, as we have found experimentally there is a considerable drop in performance when sharing also the BN parameters.


Since we can only recur over convolutional layers with the same spatial size, we add one extra convolution between recurrent residual blocks with different spatial sizes.
During backpropagation, each shared residual mapping function $\hat{f}_i(\cdot)$ as given in~\eq{shared},  
at each time $t$, will give an associated gradient for the weights to be learned in that layer, $\partial\mathbf{W}_i^s (t)$. 
As in~\cite{liang2015recurrent}, we update the shared weights of the student network in one layer, $\mathbf{W}_i^s$, 
with the sum of all the gradients across all time steps, $t$.

\begin{equation}
  \frac{ \partial{\mathcal{L}(\mathbf{x}, \{\mathbf{W}^s\})} }{ \partial{\mathbf{W}_i^s} } = \sum_{t=1}^n \frac{ \partial{\mathcal{L}(\mathbf{x}, \{\mathbf{W}^s\})} }{ \partial{\mathbf{W}_i^s} (t) },
\end{equation}
where $\mathcal{L}(\mathbf{x}, \{\mathbf{W}^s\})$ is our final loss as given in~\eq{loss}, and $n$ is the number of recurrence steps for that recurrent BRC unit.

\section{Experimental evaluation}\label{sec:experiments}
\subsection{Experimental analysis}
For all experiments we standardize the data by translating it to mean zero and scaling it to unit standard deviation. We use a weight decay of $1.0e-5$ and momentum of $0.9$ and initialize the weights following~\cite{he2015delving}.\\

\noindent \textbf{CIFAR-10~\cite{krizhevsky2009learning}.} 
We first evaluate our model on CIFAR-10 using the WRN-18-2 model from~\cite{zagoruyko2016wide} with a wide residual teacher model with 18 layers and 3 blocks
and make the student networks half as wide by using only half of the number of filters in the convolutional layers.
The teacher and student models are trained from scratch using batch size 128. 
The starting learning rate is $0.1$ and is decreased by a factor of $10$ every $40 K$ iterations and $60 K$ iterations.
For data augmentation, following \cite{he2016deep} we pad 4 pixels on each side, take random crops and add random horizontal flips. \\

\begin{table}
	\centering
	\begin{tabular}{llll} \toprule
		\# Recurs						& \multicolumn{2}{l}{Parameters} 	& Accuracy \\ 
						& \%  & \# & \\ \midrule
		Teacher			& 100\% & 1.235 M			& 93.25\% \\ \midrule						
		1				& 6\% & 73 K				& 86.95\% \\
		2				& 6\% & 73 K				& 88.17\% \\
		3				& 6\% & 73 K				& 88.33\% \\
		4				& 6\% & 74 K				& 88.39\% \\
		5				& 6\% & 74 K				& 88.24\% \\
		6				& 6\% & 74 K				& 87.95\% \\
		12				& 6\% & 75 K				& 88.03\% \\ 
 \bottomrule
	\end{tabular}
	\caption{ \small
		\textbf{Exp. 1:} 
		The effect of recurrence on performance on CIFAR-10. We test between 1 and 6 repetitions, and 12 repetitions on our lightest student model: \textsc{ReResNet-3}. 
		After 3-4 recurrences performance is stable. 
	}
	\label{tab:repetitions}
\end{table}

\noindent \textbf{Exp. 1: How many times to recur?} In table~\ref{tab:repetitions} we show the effect of recurrence on accuracy using our smallest model, \textsc{ReResNet-3}.  
For this experiment we do not use the teacher-student similarity loss. 
Performance is stable to recurrence. 
In our subsequent experiments we evaluate our models when using 3 and 6 repetitions in the recurrent links.\\

\begin{table}
	\centering
	\begin{tabular}{lllll} \toprule
		Network								& \# Recurs	& \multicolumn{2}{l}{Parameters} & Accuracy \\ 
						& & \%  & \# & \\ \midrule		
		Teacher							 	&					& 100\% & 1.235 M	& 93.28\% \\ \midrule
		\textsc{ReResNet-1}					&	3				& 10\% & 122 K			& 89.81\% \\
		\textsc{ReResNet-2}					&	3				& 10\% & 122 K			& 89.25\% \\
		\textsc{ReResNet-3}					&	3				& 6\% & 73 K			& 88.33\% \\  \midrule 
		\textsc{ReResNet-1}					&	6				& 10\% & 124 K			& 89.99\% \\
		\textsc{ReResNet-2}					&	6				& 10\% & 124 K			& 89.00\% \\
		\textsc{ReResNet-3}					&	6				& 6\% & 74 K			& 87.95\% \\
		\bottomrule
	\end{tabular}
	\caption{ \small
		\textbf{Exp. 2:} 
		Comparison of \textsc{ReResNet-1}, \textsc{ReResNet-2} and \textsc{ReResNet-3} on CIFAR-10 with 3 and 6 recurrences. 
		Overall, \textsc{ReResNet-1} performs slightly better than the other two variants, however there is not a significant difference in the performance of the three considered models.
	}
	\label{tab:model}
\end{table}

\noindent \textbf{Exp. 2: How to add recurrence?} 
In table~\ref{tab:model} we compare our three proposed variants \textsc{ReResNet-1}, \textsc{ReResNet-2}, \textsc{ReResNet-3} on CIFAR-10 when using 3 and 6 recurrences.
Here we do not use the teacher-student similarity loss. 
The \textsc{ReResNet-1} performs slightly better than the other two variants while having an intuitive manner of adding recurrence: 
the complete block is recurred, rather than each convolution in the block separately. 
In the subsequent experiments we use the \textsc{ReResNet-1} model and refer to it as \textsc{ReResNet}.\\

\noindent \textbf{Exp. 3: Comparison with existing work.} 
Table~\ref{tab:comparison} shows the accuracy and number of parameters of our \textsc{ReResNet$^*$}, enhanced with the teacher-student similarity loss, 
when compared to popular network architectures. 

We compare with Circulant CNN~\cite{cheng2015exploration} which has a similar size with our model, as well as the FitNet proposed in \cite{romero2015fitnets}.
We additionally compare with other larger models  such as NIN (Network in Network) \cite{lin2013network},
Highway networks \cite{srivastava2015highway}, Maxout network\cite{goodfellow2013maxout}, and probabilistic Maxout network \cite{springenberg2013improving}.
Our proposed approach outperforms 3 out of these 6 models considered.
We outperform the Circulant CNN \cite{cheng2015exploration} while having a similar number of parameters, 
and the two Maxout models \cite{goodfellow2013maxout,springenberg2013improving} while using 40$\times$ less parameters.

\begin{table}
	\centering
	\begin{tabular}{llll} \toprule
		Network											& \multicolumn{2}{l}{Parameters} & Accuracy \\ 
						&  \%  & \# & \\ \midrule				
		Teacher											& 100\% & 1.235 M	& 93.28\% \\ \midrule
		Maxout~\cite{goodfellow2013maxout}				& $>$400\% & $>$5 M		& 90.62\%\\
		Prob. maxout~\cite{springenberg2013improving}		& $>$400\% & $>$5 M		& 90.61\%\\
		FitNet~\cite{romero2015fitnets}					& 200\% & 2.5 M			& 91.61\%\\
		Highway~\cite{srivastava2015highway}				& 100\% & 1.25 M		& 91.20\%\\
		NIN~\cite{lin2013network}						& 78\% & 970 K			& 91.19\% \\
		Circulant CNN~\cite{cheng2015exploration}			& 9.7\% & 120 K			& 84.29\% \\ \midrule
		\textsc{ReResNet$^*$}	(3 recurs)				& 10\% & 122 K			& 90.29\%\\
		\textsc{ReResNet$^*$}	(6 recurs)				& 10\% & 124 K			& 90.49\%\\ 
		\bottomrule         
	\end{tabular}
	\caption{ \small
		\textbf{Exp. 3:}
		Comparison of performance and number of parameters on CIFAR-10 for our \textsc{ReResNet$^*$}, enhanced with the teacher-student similarity loss, 
        when compared to other efficient methods such as FitNets \cite{romero2015fitnets} 
		and Circulant CNN \cite{cheng2015exploration}, and other large architectures such as NIN (Network in network) \cite{lin2013network}, Highway networks \cite{srivastava2015highway},
		Maxout networks \cite{goodfellow2013maxout}, and probabilistic Maxout networks \cite{springenberg2013improving}. 
		Our method outperforms the Circulant CNN while having a similar number of parameters, and performs on par with the Maxout architectures \cite{goodfellow2013maxout,springenberg2013improving} 
		while having 40$\times$ less parameters.
	}
	\label{tab:comparison}
\end{table}

\subsection{Performance on scene recognition} 
To evaluate the generalization capacity of our method we evaluate the performance of our \textsc{ReResNet$^*$} enhanced with the teacher-student similarity loss, 
on two scene datasets.\\

\noindent \textbf{Exp. 4: Scenes~\cite{quattoni2009recognizing}.} Given the small dataset we use a shallower but wider teacher model: 
a wide residual net from~\cite{zagoruyko2016wide} with width factor 4 and 12 layers, using only 2 residual blocks rather than 3. 
Given the small dataset size, we extract features from the last residual group of a pre-trained ResNet-50~\cite{he2016deep} on ImageNet~\cite{deng2009imagenet} 
and train our teacher\slash student models on top of that. 
Batch size is set to 128. 
The learning rates used are $0.01$ for the first $2 K$ iterations and $0.001$ for the rest. 
Following~\cite{mahmood2016resfeats}, we resize the input images to 224$\times$224 pixels. Table~\ref{tab:scenes} shows the performance on the Scenes dataset when comparing our teacher performance with the performance of our \textsc{ReResNet$^*$} student architecture. 
We notice a 3\% decrease in performance at the gain of more than $5 \times$ reduction in parameters.\\

\noindent \textbf{Exp. 5: MiniPlaces~\cite{zhou2017places}.} On MiniPlaces the teacher architecture corresponds to the standard residual network proposed in \cite{he2016deep}, ResNet-34. 
Here the student model is as wide as the teacher model. 
The teacher and student networks are trained from scratch using batch size 256. 
We start with learning rate $0.1$, and we decrease it by 10 every $70 K$ and $160 K$ iterations. 
As input during training we use random image crops of 112$\times$112 pixels from the initial 128$\times$ 128 images. In table~\ref{tab:miniplaces} the performance on MiniPlaces is evaluated, where we compare the teacher performance with the performance of our \textsc{ReResNet$^*$} student model.
In this experiment we obtain the same performance as our teacher model while using $3 \times$ less parameters in our student model.

\begin{table}
\centering
	\begin{tabular}{llll} \toprule
	Network					& \multicolumn{2}{l}{Parameters}		& Accuracy\\ 
						    &  \%  & \# & \\ \midrule					
	Teacher					& 100\% & 23.64 M			& 71.80\% \\ \midrule
	\textsc{ReResNet$^*$}		& 18\% & 4.15 M			    & 68.73\% \\ \bottomrule
	\end{tabular} 
\caption{ \small
	\textbf{Exp. 4:}
	On the Scenes dataset we lose 3\% in accuracy at for a more than 5 $\times$ gain in parameter usage. 
}
\label{tab:scenes}
\end{table}

\begin{table}
\centering
	\begin{tabular}{lllll} \toprule	
	Network				& \multicolumn{2}{l}{Parameters}		& Top-1				& Top-5 \\ 
						&  \%  & \# & & \\ \midrule					
	Teacher				& 100\% & 6.07 M			& 47.54\%			& 76.82\% \\ \midrule
	\textsc{ReResNet$^*$}	& 33\% & 1.98 M			& 47.56\%	& 77.42\% \\ \bottomrule       
	\end{tabular}
\caption{ \small
	\textbf{Exp. 5:}
	On Miniplaces the student performs similar to the teacher while using 3$\times$ less parameters. 
}
\label{tab:miniplaces}
\end{table}

\section{Conclusion}\label{sec:conclusion}
In this paper we focus on model parameter reduction by using knowledge distillation for learning compact student models from wider and deeper residual models.
Our student models use recurrent connections for compacting the information and allowing for shallow student networks. 
We propose three variants of our student model, \textsc{ReResNet-1}, \textsc{ReResNet-2} and \textsc{ReResNet-3}, 
in which the convolutional layers in the residual blocks are tied over time in different manners. 
We evaluate our model choices as well as compare with existing work in terms of accuracy and used number of parameters, 
and show experimentally that our models can achieve comparable performance, using considerably less parameters. 

One limitation of our approach is that recurrence is now added simply through sharing parameters over time which lacks in descriptive power.
We believe improvements can be obtained by using gating functions as in the case of LSTM~\cite{hochreiter1997long} blocks, to control the information remembered. 
The student networks would benefit from this approach.

\bibliographystyle{IEEEbib}
{\small
\bibliography{pintea}
}
\end{document}